\documentclass{article}%
\usepackage{amsmath}%
\usepackage{amsfonts}%
\usepackage{amssymb}%
\usepackage{graphicx}

\begin{document}

\author{G.E. Miram and V.K. Petrov. \thanks{Institute for Theoretical Physics,
National Academy of Sciences of Ukraine}}
\title{{\Large A Probabilistic Model of Machine Translation }}
\date{}
\maketitle

\begin{abstract}
A probabilistic model for computer-based generation of a machine translation
system on the basis of English-Russian parallel text corpora is suggested. The
model is trained using parallel text corpora with pre-aligned source and
target sentences. The training of the model results in a bilingual dictionary
of words and \textquotedblright word blocks\textquotedblright\ with relevant
translation probability

\end{abstract}

\section{Introduction.}

The corpus-based statistical MT gains more popularity nowadays due to vastly
increased capacity of modern computers. The works of P. Brown and
collaborators \cite{Brown+al:statistical,Brown+al:math}, may be regarded as a
typical recent example.

This paper suggests another approach to statistical MT different from that of
Brown et al. The suggested model is trained on pre-aligned bilingual text
corpora and the following approach to 'tailor making' a computer dictionary
and an MT system is taken. The translation of a source word combination by a
target one is determined by the correlation with the neighboring word
combinations both in the source and the target texts rather than only by the
translation probabilities of the combinations themselves.

The word order of the source and target sentences seldom coincide, however,
the raw translation with the incorrect order of words may often be understood
by a specialist. The translation quality will radically improve if instead of
individual words one takes internally agreed word combinations with fixed
order (blocks).

In this model statistically stable source blocks are related to the most
probable target ones using specially introduced function, called ''adhesion
function'' since it is believed that this function indirectly reflects the
grammatical and semantic ''adhesion'' of the words in a text. We believe that
blocks with negative correlation having been excluded the remaining internally
agreed blocks in a way will become a substitute of the proper word order in
the target sentence.

\section{ Probability Assessment and Model Training.}

The training corpus is presumed to be pre-aligned, i.e. divided into the
matching pairs of the source and target sentences\footnote{In the training
corpus as well as in the translated texts the ends of sentences are presumed
to be marked with common punctuation marks.}. Each of the sentences comprising
a pair is broken into the word sequences (blocks) in such a way that the order
of words a sequence had in the sentence is preserved.

The first block comprises the first words of matching sentences, then one word
is added each time until the block reaches the extreme length\footnote{In our
case the minimal length limit of a block is three words.}. Then the procedure
is repeated starting from the second word and so on. All the blocks obtained
in the above manner are stored in temporary data file with the blocks that
appear several times being regarded at this stage as different (viz. Fig. 1).

\begin{center}%
\begin{table}[tbp] \centering
\begin{tabular}
[c]{|cccccccc}\hline\hline
\multicolumn{1}{||c}{{\small John}} & {\small and} & {\small Mary} &
{\small go} & {\small to} & {\small school} & {\small every} &
\multicolumn{1}{c||}{{\small morning}}\\\hline\hline
$\Downarrow$ & $\Downarrow$ & $\Downarrow$ & $\Downarrow$ & $\Downarrow$ &
$\Downarrow$ & $\Downarrow$ & \multicolumn{1}{c|}{$\Downarrow$}\\\cline{2-4}
& \multicolumn{1}{||c}{{\small and}} & {\small Mary} & {\small go} &
\multicolumn{1}{||c}{} &  &  & \\\cline{2-6}\cline{2-6}
&  & \multicolumn{1}{||c}{{\small Mary}} & {\small go} & {\small to} &
{\small school} & \multicolumn{1}{||c}{} & \\\hline
\multicolumn{1}{||c}{{\small John}} & {\small and} & {\small Mary} &
\multicolumn{1}{||c}{} &  &  & \multicolumn{1}{||c}{{\small every}} &
\multicolumn{1}{c||}{{\small morning}}\\\cline{1-3}\cline{7-8}%
\end{tabular}%
\end{table}%

Fig.1. Diagram illustrating the breaking of the sentence into blocks.
\end{center}

$\mathbf{\qquad}$We suggest two alternative procedures for the sorting-out of
the preliminary data file to obtain the translation dictionary.

Let in a sentence of a length $L$ number of $b-$ words blocks is $L-b+1.$Total
number of blocks with the length that does not exceed $l$ in this sentence is
\begin{equation}
N_{l}=\sum_{b=1}^{l}\left(  L-b+1\right)  =\frac{l\left(  2L-l+1\right)  }%
{2}\leq N_{L}=\frac{L\left(  L+1\right)  }{2}\label{numb}%
\end{equation}

Number of block pairs
\begin{equation}
N_{l}^{\left(  S\right)  }\times N_{l}^{\left(  T\right)  }=\frac{l^{2}\left(
2L^{\left(  S\right)  }-l+1\right)  \left(  2L^{\left(  T\right)
}-l+1\right)  }4\label{corresp}%
\end{equation}

where $L^{\left(  S\right)  }$ and $L^{\left(  T\right)  }$ are lengths of
source and target sentences correspondingly.

Even in a texts where lengths of source and target sentences are large enough
(say $L^{\left(  S\right)  }$ $=L^{\left(  T\right)  }=20$ )

\begin{center}%
\begin{table}[tbp] \centering
\begin{tabular}
[c]{||l||l||l||l||l||}\hline\hline
$N_{l}^{\left(  S\right)  }=N_{20}^{\left(  T\right)  }$ & $N_{3}^{\left(
S\right)  }=N_{3}^{\left(  T\right)  }$ & $N_{20}^{\left(  S\right)  }\times
N_{20}^{\left(  T\right)  }$ & $N_{3}^{\left(  S\right)  }\times
N_{3}^{\left(  T\right)  }$ & $\frac{N_{20}^{\left(  S\right)  }\times
N_{20}^{\left(  T\right)  }}{N_{3}^{\left(  S\right)  }\times N_{3}^{\left(
T\right)  }}$\\\hline\hline
$420$ & $114$ & $176400$ & $12996$ & $13573$\\\hline\hline
\end{tabular}
Table1
\end{table}%

\end{center}

We see that whole volume of block pairs less then $14$ times larger then the
number of block pairs with the length that does not exceed $l$ $=3$ .

\subsubsection{a) ''All-In'' Relations Alternative.}

\begin{enumerate}
\item In this case the sentence pairs are broken into arbitrary number $v$ and
$\widetilde{v}$ of the blocks.

\item If the number of different blocks, $v$, obtained after all possible
divisions of a sentence is greater than the number of blocks, $\widetilde{v}$,
obtained in its counterpart, the latter is added with blank blocks until the
number of blocks in both target and source sentences becomes equal
$w=\max\left(  v;\widetilde{v}\right)  .$

\item Let us relate each block of the source (English) sentence with its all
target (Russian) counterparts

\item The resulting $w^{2}$ pairs $\left\{  S_{j,}T_{k}\right\}  ;$
$j;k=1,2,...,w$ ; are stored in the temporary data file.
\end{enumerate}

\subsubsection{b) Symmetrical Relations Alternative.}

\begin{enumerate}
\item In this case the sentence pairs are broken into the equal number $w$ of
the blocks having no blank counterparts. Moreover, only the blocks with the
same value of $j$ are stored in the preliminary data file.

\item Then for each division we shall have $w$ pairs $\left\{  S_{j,}%
T_{j}\right\}  ;$ $j=1,2,...,w$.

\item The resulting $w$ pairs $\left\{  S_{j,}T_{k}\right\}  ;$ $j,k=1,2,...,w
$ are stored in the temporary data file.
\end{enumerate}

The symmetrical alternative will require a bigger training corpus, however, it
will allow to use the same block comparison procedure in training and in translation.

In both alternatives the procedure of sentence division will be terminated
when the computer storage capacity is exhausted.

Let then $n={\sum_{s}}w_{s}^{2}$ be the total number of the matching block
pairs of the alternative \textquotedblright a\textquotedblright\ $n={\sum_{s}%
}w_{s}$ be that of the alternative \textquotedblright b\textquotedblright.
Then the total number of the source blocks $S$ $=\left(  s_{1},s_{2}%
,...\right)  $, that of the target blocks $T$ $=\left(  t_{1},t_{2}%
,...\right)  $ and the total of the pairs $\left\{  S,T\right\}  $ will be
given by $n^{S}$, $n^{T}$, $n^{S^{\bigcap}T}$, whereas the relevant
probabilities $P^{S}$, $P^{T}$ and $P^{S^{\bigcap}T}$will be found from:
\begin{equation}
P^{T}=\frac{n^{T}}{n};\qquad P^{S}=\frac{n^{S}}{n};\qquad P^{T^{\bigcap}%
S}=\frac{n^{T^{\bigcap}S}}{n}\label{probab1}%
\end{equation}
Then the conventional probability $P\left(  T|S\right)  =$ $P^{T^{\bigcap}%
S}/P^{S}$ of the word $T$ as a translation of the word $S$ and the probability
$P\left(  S|T\right)  =$ $P^{T^{\bigcap}S}/P^{T}$ of the word as a translation
of the word are related by Bayes formula
\begin{equation}
P\left(  T|S\right)  P^{S}=P^{T^{\bigcap}S}=P\left(  S|T\right)
P^{T}\label{cond2}%
\end{equation}

As it is well known, the correlation between the events and will be
\begin{equation}
C^{T^{\bigcap}S}=P^{T^{\bigcap}S}-P^{S}P^{T}\label{corr3}%
\end{equation}
When events $S$ and $T$ are independent, i.e. co-occur at random,
$P^{T^{\bigcap}S}=P^{S}P^{T}$ , the correlation function $C^{T^{\bigcap}S}$
becomes zero. Then it may be suggested that the negative correlation
$C^{T^{\bigcap}S}$ $<0$ will be the case, when the source and target language
words are so eager to avoid each other, that their correlated co-occurrence is
less probable than the random one. We regard such co-occurrences as prohibited
by the rules of the languages involved.

The correlation analysis starts from the minimal, one-word blocks $S=\left(
s_{1}\right)  ,T=\left(  t_{1}\right)  $. The longer two-word $S=\left(
s_{1},s_{2}\right)  $, $T=\left(  t_{1},t_{2}\right)  $ and three-word
$T=\left(  t_{1},t_{2},t_{3}\right)  $, $S=\left(  s_{1},s_{2},s_{3}\right)  $
ones are analysed if%

\begin{equation}
C^{\left(  s_{1},s_{2}\right)  }=P^{\left(  s_{1},s_{2}\right)  }-P^{\left(
s_{1}\right)  }P^{\left(  s_{2}\right)  }>0
\end{equation}
and accordingly if%

\begin{equation}
P^{\left(  t_{1},t_{2},t_{3}\right)  }-P^{\left(  t_{1},t_{2}\right)
}P^{\left(  t_{3}\right)  }>0
\end{equation}
or%

\begin{equation}
P^{\left(  t_{1},t_{2},t_{3}\right)  }-P^{\left(  t_{1}\right)  }P^{\left(
,t_{2},t_{3}\right)  }>0
\end{equation}
and so on.

To save the storage space we shall pay attention only to closely correlated
events, for which the relative correlation\footnote{Absolute correlation value
of the blocks divided by the value of their random co-occurrence which
accounts for the rare but strongly correlated events.} :
\begin{equation}
\rho^{S_{j}\bigcap S_{j+1}}=\frac{C^{S_{j}^{\bigcap}S_{j+1}}}{P^{S_{j}%
}P^{S_{j+1}}};\ \rho^{T^{\bigcap}S}=\frac{C^{T^{\bigcap}S}}{P^{S_{j}}P^{T}%
}\label{rho4}%
\end{equation}
to satisfy the condition
\begin{equation}
\rho^{S_{j}\bigcap S_{j+1}}>c_{+}^{S};\qquad\rho^{S_{j}\bigcap S_{j+1}}%
<-c_{-}^{S}\label{restrA}%
\end{equation}

\begin{equation}
\rho^{T_{j}\bigcap T_{j+1}}>c_{+}^{T};\ \rho^{T_{j}\bigcap T_{j+1}}<-c_{-}%
^{T}\label{restrB}%
\end{equation}
\begin{equation}
\rho^{T_{j}\bigcap S_{k}}>c_{+}^{TS};\ \rho^{T_{j}\bigcap S_{k}}<-c_{-}%
^{TS}\label{restrC}%
\end{equation}

All pairs $S_{j}\bigcup S_{j+1},$consisting of sub-blocks $S_{j}$ and
$S_{j+1}$ will be included in $S$-dictionary, if sub-blocks $S_{j}$ and
$S_{j+1}$ satisfy the condition $\left(  \text{\ref{restrA}}\right)  $.
Similarly if sub-blocks $T_{j}$ and $T_{j+1}$satisfy the condition$\left(
\text{\ref{restrB}}\right)  $ they are included in a $T$-dictionary. And,
finally, if the condition $\left(  \text{\ref{restrC}}\right)  $ is satisfied,
we include $\left(  T_{j},S_{k}\right)  $pairs into $TS$-dictionary. The
values of positive constants $c_{\pm}^{S}$, $c_{\pm}^{T}$and $c_{\pm}^{TS} $
naturally depend on the computer storage capacity. In this way we shall be
able to calculate both $P\left(  S|T\right)  $ and $P\left(  T|S\right)  $
which will allow to reverse the direction of the translation.

All the elements of a word paradigm enter the dictionaries as separate
entries. Both the selection of a correct (and strongly prohibited) form for
translation and agreement between the forms are achievable, on the one hand,
because the forms within a block are already agreed and, on the other, because
reasonable agreement of paradigm forms in matching blocks is obtained in the
course of maximisation, as described below.

The training may be simplified if we have a dictionary of
cognates\footnote{The cognates are the words of similar graphic image in
different languages, e. g. syntax and sintaksis.} . In this case the
preliminary data file will not include the pairs in which one block comprises
a cognate whereas its counterpart does not\footnote{Identification and use of
cognates may be found, e.g., in \cite{Church:program},and
\cite{Miram:approach}.} When the dictionary is generated (i.e. available
amount of training corpora is exhausted), we pass over to the translation
using a new text.

\section{ Translation Model Optimisation}

The translation of a new sentence starts from dividing it into blocks. This is
being done in such a way that none of the blocks is wholly contained in any
other. To satisfy this condition any next block will begin with, at least, one
word after the first word of the previous block and will end with, at least,
one word after the last word of the preceding block. Each of the source blocks
will be related to the target ones.

The division starts from the blocks of the maximum length available in the
dictionary, and the block length is gradually decreased to the word-to-word
pairs. To select the optimal translations we shall use the following
maximisation procedure.

For the words in a source (or target) text we suggest the characteristic of
'adhesion''. We shall call ''adhered'' both the words which enter one and the
same block and those entering the overlapping blocks. Thus, in Fig. 1 the
words $abc$, $bcd$, $cdef$ and $gh$ adhere into blocks and since the words
$b$, $c$, $d$ enter several blocks simultaneously they are also considered
adhered. Words $ag$, $ah$, $bg$, $bh$ and so forth are not adhered. Fig 1.
shows the source sentence only. It is understood that for simplicity the
target sentence will have the same block pattern. Naturally, in both texts the
blocks with multiple overlapping will be those having greater adhesion. At the
same time, the longer is the block the smaller is its occurrence probability
in the dictionary after training . For equal competition opportunities for
longer and shorter blocks the following procedure is suggested. To illustrate
this let us consider the blocks of maximum two words and assume that a
three-word sentence is translated by the two linked blocks (Fig. 2 ).

\begin{center}%
\begin{table}[tbp] \centering
\begin{tabular}
[c]{||l||l||l||l||l||l||l||}\hline\hline
$\left[  s_{1}s_{2}s_{3}\right]  $ & $\approx$ & ${\left[  s_{1}%
\widetilde{s_{2}}\right]  }{\left[  \widetilde{s}_{2}s_{3}\right]  }$ &
$\Longrightarrow$translation$\Longrightarrow$ & ${\left[  t_{1}\widetilde
{t_{2}}\right]  }{\left[  \widetilde{t_{2}}t_{3}\right]  }$ & $\approx$ &
$\left[  t_{1}t_{2}t_{3}\right]  $\\\hline\hline
\end{tabular}
%

Fig. 2. A Diagram of a 3-Word Sentence Translated by Two Overlapping Blocks.
\end{table}%

\end{center}

Of course, all the words in Fig. 2 are adhered and the source sentence cannot
be translated by one target sentence only because of our two-word block
constraints. We suggest the following model-type relation to compute the true
probability :
\begin{equation}
P^{\left(  t_{1},t_{2},t_{3}\right)  }\approx P^{\left(  t_{1},t_{2}\right)
}P^{\left(  t_{2},t_{3}\right)  }f\left(  P\left(  t_{2}\right)  \right)
\label{model6}%
\end{equation}
i.e. we suggest that the relation of the true probability to the probabilities
of the individual blocks, $P^{\left(  t_{1},t_{2}\right)  }$and, $P^{\left(
t_{2},t_{3}\right)  }$depends only on the probability of the overlapping words
$P^{\left(  t_{2}\right)  }$\footnote{For the sake of simplicity we show the
adhesion function only for the target blocks, it is understood, however, that
similar function is calculated in the same way for the source blocks as
well.}. Generally speaking, finding the overlapping probability function
$f\left(  P\left(  t_{2}\right)  \right)  $ requires a special
phenomenological study, but for our model we limit ourselves with the
following simple considerations. It is easy to see that if all the words are
not adhered with the others, then%

\begin{equation}
P^{\left(  t_{1},t_{2},t_{3}\right)  }\approx P^{\left(  t_{1}\right)
}\left[  P^{\left(  t_{2}\right)  }\right]  ^{2}P^{\left(  t_{3}\right)
}\label{indep7}%
\end{equation}
and hence having substituted $\left(  \ref{indep7}\right)  $ into $\left(
\ref{model6}\right)  $ we obtain for this very special case:
\begin{equation}
f\left(  P\right)  \approx1/P\label{appr8}%
\end{equation}
We hope that this approximation will give satisfactory results for the general
case, that is why we assign the factor $1/P$ each time the words in blocks overlap.

The function $f\left(  P\right)  $ is introduced to accord the blocks and its
form is presumed to be universal for the given language. We shall call it
global adhesion factor (GAF). A more effective way to account for the
overlapping of the blocks is to introduce local adhesion factor( LAF ) for
each word rather than GAF:
\begin{equation}
f_{t_{2}}=\frac{P^{\left(  t_{1},t_{2},t_{3}\right)  }}{P^{\left(  t_{1}%
,t_{2}\right)  }P^{\left(  t_{2},t_{3}\right)  }}\label{adh9}%
\end{equation}
LAF $f_{t_{2}}$ for each $t_{2}$- word is at first computed for all
$P^{\left(  t_{1},t_{2},t_{3}\right)  }$, $P^{\left(  t_{1},t_{2}\right)  }$
and $P^{\left(  t_{2},t_{3}\right)  }$ available and then averaged over
$t_{1}$and $t_{3}$. In this case $f_{t_{2}}$really becomes an inherent
characteristics of an $\widetilde{t}_{2}$ - word. It easy to see that in
$P\left(  S|T\right)  =P^{S^{\bigcap}T}/P^{T}$ the overlapping of
$\widetilde{t}$- words is present both in $P^{S^{\bigcap}T}$and $P^{T}$,
hence, LAF values for $t$- words are cancelled, then during translation stage
we take into account only LAF for $s$-words. Then for a sentence we have:
\begin{equation}
F_{L}^{S}={\prod}_{j}F_{L}^{S_{j}\bigcap S_{j+1}}={\prod}_{j}\left[  {\prod
}_{\left(  \widetilde{s}_{\mu}\in S_{j}\bigcap S_{j+1}\right)  }%
f_{\widetilde{s}_{\mu}}\right] \label{Overlapp10}%
\end{equation}
where the product is computed over overlapping $\widetilde{s}$-words.

An overlapping in the source sentences ( e.g. $s_{2}$ in Fig. 2) may be
related to that in the target sentence (e.g. $t_{2}$ in Fig. 2). During
translation combining the target blocks we may get double occurrence of the
overlapping word ( e.g., when combining $T=\left(  t_{1},t_{2}\right)  $and
$\acute{T}=\left(  t_{2},t_{3}\right)  $ we get double occurrence of $t_{2}$)
, which are to be excluded from the translation product. One should also
exclude the synonyms as well. Having excluded double occurrences we shall
obtain a set of $\mu=1,2,...$ translation alternatives $\left\{  T_{k}^{\mu
}\bigcup T_{k+1}^{\mu}\right\}  $ combining several neighboring blocks $k$ and
$k+1$, some of which may be grammatically incorrect.

We suggest the following correction procedure:

a)Each of $\left\{  T_{k}^{\mu}\bigcup T_{k+1}^{\mu}\right\}  $alternatives is
broken into all possible sub-blocks ;

b)The optimal alternative is obtained by%

\begin{equation}
\max\left\{  \rho^{T_{k}^{\mu}\bigcap T_{k+1}^{\mu}}\right\} \label{alt11}%
\end{equation}
We believe that increasing the length of blocks we shall be able to select
successfully the translation words corresponding to the source context.
Moreover, one will hardly require fragments longer than four words, since
correlation at such distances seems rather weak.

For the general case of translation probability maximisation we propose the
following:
\begin{equation}
\max\left\{  \overset{N}{{\prod}_{j}}\frac{P^{T_{j}^{\bigcap}S_{j}}}{P^{S_{j}%
}}F_{L}^{T_{j}T_{j+1}}\right\} \label{transl-12}%
\end{equation}
\noindent where $P^{T_{j}^{\bigcap}S_{j}}=P^{\left(  t_{1},t_{2},...\right)
_{j}^{\bigcap}\left(  s_{1},s_{2},...\right)  _{j}}$is the probability
corresponding to block $j$ in given translation alternative. The overlapping
function $F_{L}^{T_{j}T_{j+1}}$ for $n_{\mu}$-fold overlapping of the words
$\widetilde{t}_{\mu}$ in a neighbored blocks $T_{j}$ and $T_{j+1}$may be
computed as $F_{L}^{T_{j}T_{j+1}}=\left\{  {{\prod}_{\mu}P^{\left(
\widetilde{t}_{\mu}\right)  }}\right\}  _{\widetilde{t}_{\mu}\in
T_{j}^{\bigcap}T_{j+1}}$ (see $\left(  \text{\ref{Overlapp10}}\right)  $). The
maximisation procedure can be easily modified for the source language since
the suggested model is evidently symmetrical.

\section{ Conclusions}

Similar to \cite{Brown+al:statistical} we train our model using parallel text
corpora. However, our model is different in a number of aspects. We consider
the suggested numerical correlation between source and target blocks
(simultaneous interpreter principle) more critical for translation quality
than selection of optimal word positions through the maximisation of the
product of the relevant probabilities as in\cite{Brown+al:statistical}
\cite{Brown+al:math}. For the model, suggested in this paper, there is room
for perfection limited only by computation capacity through increasing the
block length. In the model of \cite{Brown+al:statistical},
\cite{Brown+al:math}, however, it is not clear, how without some new modelling
ideas to make probability-based choice between, say, such two sentences as
''He is alive, but she is dead'' and ''He is dead, but she is alive'' both of
which are correct grammatically, but controversial semantically.

\end{document}